\documentclass{article}
\usepackage{spconf,amsmath,graphicx}

\usepackage{enumitem}
\setlist{nosep, leftmargin=14pt}

\usepackage{mwe} 
\usepackage{booktabs}

\title{ENHANCING ZERO-SHOT LEARNING IN MEDICAL IMAGING: INTEGRATING CLIP WITH ADVANCED TECHNIQUES FOR IMPROVED CHEST X-RAY ANALYSIS}
\name{Prakhar Bhardwaj, Sheethal Bhat, Andreas Maier}
\address{Fakultät für Pattern Recognition, FAU Erlangen-Nürnberg, Germany}
\begin{document}
%
\maketitle
\begin{abstract}
Due to the large volume of medical imaging data, advanced AI methodologies are needed to assist radiologists in diagnosing thoracic diseases from chest X-rays (CXRs). Existing deep learning models often require large, labeled datasets, which are scarce in medical imaging due to the time-consuming and expert-driven annotation process. In this paper, we extend the existing approach to enhance zero-shot learning in medical imaging by integrating Contrastive Language-Image Pre-training (CLIP) with Momentum Contrast (MoCo), resulting in our proposed model, MoCoCLIP. Our method address challenges posed by class-imbalanced and unlabeled datasets, enabling improved detection of
pulmonary pathologies. Experimental results on the NIH ChestXray14 dataset demonstrate that MoCoCLIP outperforms the state-of-the-art CheXZero model, achieving relative improvement of approximately 6.5\%. Furthermore, on the CheXpert dataset, MoCoCLIP demonstrates superior zero-shot performance, achieving an average AUC of 0.750 compared to CheXZero’s 0.746, highlighting its enhanced generalization capabilities on unseen data.
\end{abstract}
\section{Introduction}
\label{sec:intro}


Artificial Intelligence (AI) has revolutionized the field of medical imaging, offering significant potential to enhance diagnostic accuracy and efficiency.
In this regard, deep learning (DL) techniques used in analyzing Chest X-rays (CXRs), have demonstrated promising results as tools to aid clinical diagnosis. Despite these advancements, the development of highly accurate DL models is often hindered by the scarcity of large scale labeled medical datasets. Annotating medical images is a time-consuming and expensive process requiring expert knowledge. To address these limitations self-supervised learning in the medical domain have gained attention recently, enabling models to learn useful representations from unlabeled data.

Among the various self-supervised learning methods, image-text models are gaining popularity because they exploit the natural alignment between images and their textual descriptions. This approach is particularly transferable to the medical domain, especially for CXRs, because these images are typically accompanied by clinical reports. It has been shown that integrating radiology reports with CXR images enhances model performance by providing rich contextual information \cite{tiu2022ChexZero}. Radiological reports typically contain detailed descriptions and interpretations that can guide the models to learn semantically rich representations.

In this paper, we propose a method to enhance zero-shot learning in medical imaging by integrating Momentum Contrast (MoCo) \cite{MoCo} into the Contrastive Language-Image Pre-training (CLIP) framework \cite{DBLP:CLIP}. By incorporating MoCo, our method extends the standard CLIP approach by enabling the image encoder to learn more robust and discriminative features through momentum-based contrastive learning. This results in richer visual representations that are better aligned with textual embeddings derived from radiology reports. We demonstrate that our method outperforms the state-of-the-art CheXZero model on two distinct public datasets, NIH CXR14 \cite{NIHdata} and CheXpert \cite{irvin2019chexpert}, establishing a new benchmark in zero-shot classification of chest X-rays. Additional ablation studies evaluate the impact of various approaches that contribute to the improved performance of our proposed model. 

\section{Dataset}
\label{sec:dataset}

The NIH CXR14 Dataset \cite{NIHdata}, comprises of 112,120 X-ray images from 30,805 unique patients. The images are annotated with 14 disease categories and a “No Finding” class. The dataset was divided in 80:10:10 ratio for training, validation and testing. This split ensures a sufficiently large training set of approximately 90,000 frontal view images to train models with reasonable performance.

For zero-shot inference, we used the CheXpert test dataset \cite{irvin2019chexpert}, a benchmark widely accepted by the medical imaging community for CXR interpretation tasks. The test dataset contains 500 frontal CXR images, each labeled for the presence of 14 conditions, including \emph{Atelectasis}, \emph{Cardiomegaly}, \emph{Consolidation}, \emph{Edema}, and \emph{Pneumonia}. This dataset allows us to evaluate how well the models generalize to pathologies not explicitly observed during training.

\section{Methods}
\label{sec:pagestyle}

\subsection{Baseline Model: CLIP with CheXZero Pre-trained weights}

The baseline model is based on the CLIP framework initialised with pretrained weights from CheXZero \cite{tiu2022ChexZero}, utilizing CLIP for zero-shot medical image classification. The model comprises an image encoder (ViT-B/32) and a text encoder, both initialized with weights pre-trained on the MIMIC-CXR dataset \cite{MIMIC}. The image encoder processes images resized to $224 \times 224$ pixels, while the text encoder handles tokenized text reports with a maximum length of 77 tokens.

\subsection{Integration of Momentum Contrast (MoCo)}
\label{ssec:moco_integration}

To enhance the model’s ability to learn robust representations from unlabeled data, we integrated MoCo \cite{MoCo} into the image encoder of the CLIP model. MoCo maintains a dynamic dictionary of encoded features using a queue and a momentum-updated encoder, enabling effective contrastive learning with small batch sizes.
\begin{figure}[t]
  \centering
  \centerline{\includegraphics[width=8.0cm]{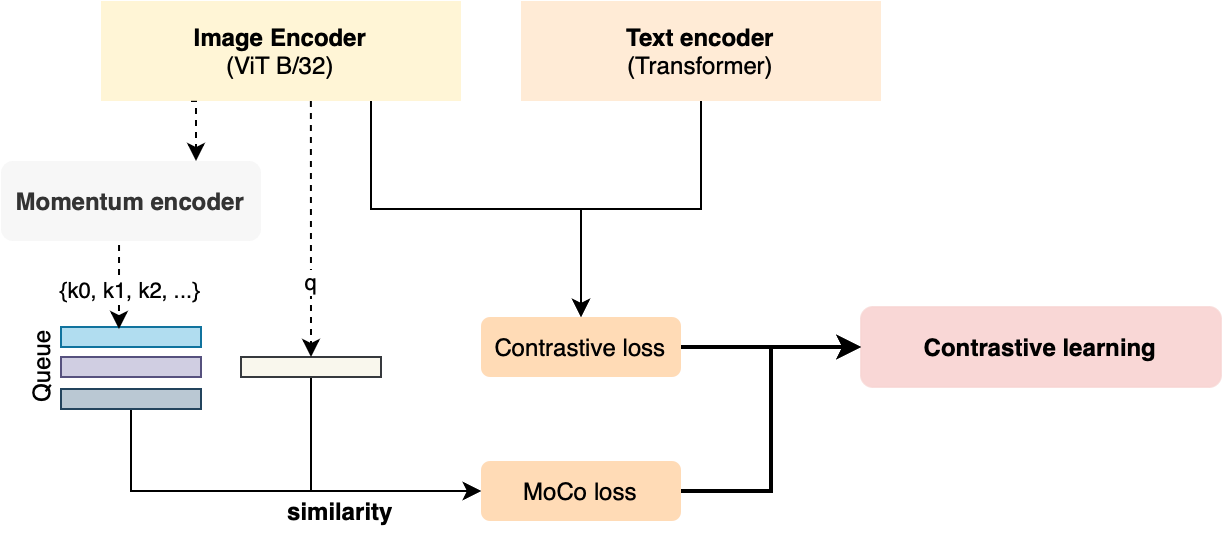}}
  \centerline{}\medskip
\caption{Training pipeline with MoCo integration into the baseline CLIP.}
\label{fig:moco}
\end{figure}

As shown in Figure \ref{fig:moco}, key modifications include:

\begin{itemize}
\item Maintaining two copies of the image encoder: the main encoder ($\theta_q$) and the momentum encoder ($\theta_k$).
\item Updating $\theta_k$ using a momentum update: $\theta_k \leftarrow m\theta_k + (1 - m)\theta_q$, where $m$ is the momentum coefficient.
\item Initializing a queue to store encoded features (keys) with a size of 65,536 and feature dimension matching the encoder’s output.
\item Using the InfoNCE loss function to compute contrastive loss between queries and keys.
\end{itemize}

\subsection{Loss Function and Optimization}

The overall loss function $\mathcal{L}$ comprises two components:

\begin{equation}
\mathcal{L} = \mathcal{L}_{\text{con}} + \lambda \mathcal{L}_{\text{MoCo}}
\end{equation}

where $\mathcal{L}_{\text{con}}$ is the standard image-text contrastive loss from CLIP, $\mathcal{L}_{\text{MoCo}}$ is the MoCo loss, and $\lambda$ is a weighting factor set to 1.0.

\subsection{Experimental Setup}

All images were preprocessed by resizing to $224 \times 224$ pixels and normalized using typical ImageNet mean and standard deviation values. Based on the content and structure of clinical reports, synthetic text reports were created for each image by utilizing the available labels. An example template for \emph{Cardiomegaly} is: \emph{Enlargement of the heart shadow is evident.} We used the Adam optimizer with an initial learning rate of $1 \times 10^{-4}$ and weight decay of $1 \times 10^{-4}$. The momentum coefficient $m$ was set to 0.999. The model was implemented using PyTorch and trained on NVIDIA Tesla V100 GPUs.

\subsection{Zero-Shot Inference}

Zero-shot classification is performed by computing similarities between the image embeddings and text embeddings of predefined prompts for each pathology. The probability of each class is calculated using a softmax over these similarities.

\section{Results and Discussions}
\label{sec:typestyle}
In this section, we present our experimental results on the NIH CXR14 dataset \cite{NIHdata} and the CheXpert test dataset \cite{irvin2019chexpert}. We compare the performance of our proposed model, \textbf{MoCoCLIP}, with baseline models to evaluate the effectiveness of integrating MoCo into the CLIP framework for zero-shot chest X-ray classification.

\subsection{Zero-Shot Inference on NIH CXR14 Dataset}
\label{ssec:subhead}
To establish a baseline, we first evaluated the \textbf{CheXZero} model \cite{tiu2022ChexZero}, which is pre-trained on the MIMIC-CXR dataset, on the NIH test set. CheXZero achieved an average AUC of 0.677 across all pathologies. It performed well on certain conditions such as Edema (AUC = 0.880), Cardiomegaly (AUC = 0.825), and Pneumothorax (AUC = 0.764). However, lower AUC scores were observed for pathologies like No Finding (AUC = 0.277) and Infiltration (AUC = 0.642), indicating potential areas for improvement. Next, we evaluated two baseline models: \textbf{ImCLIP}- CLIP model initialized with ImageNet pre-trained weights and \textbf{CXRCLIP}- CLIP model initialized with CheXZero pre-trained weights.

The ImCLIP model achieved an average AUC of 0.556, which is significantly lower than both CheXZero and CXRCLIP. This highlights the limitations of using non-domain-specific pre-trained weights for medical imaging tasks. The poor performance of ImCLIP underscores the necessity of domain adaptation for effective medical image analysis. The CXRCLIP model showed substantial improvements over ImCLIP, achieving an average AUC of 0.720. Significant gains were observed in pathologies such as: for Cardiomegaly, AUC increased from 0.576 (ImCLIP) to 0.660 (CXRCLIP). For Effusion, AUC score increased from 0.653 to 0.850 and for Infiltration, AUC increased from 0.619 to 0.690. 
\begin{table}[t]
    \centering
    \caption{Comparison of AUC scores for various pathologies using CheXZero, ImCLIP (CLIP initialized with ImageNet weights), CXRCLIP (CLIP initialized with CheXZero weights), and MoCoCLIP (our proposed model integrating MoCo into CLIP).}
    \footnotesize  
    \begin{tabular}{lcccc}
    \toprule
    \textbf{Pathology}           & \textbf{ChexZero} & \textbf{ImCLIP} & \textbf{CXRCLIP} & \textbf{MoCoCLIP} \\
    \midrule
    
    Atelectasis                  & 0.758            & 0.484              & \textbf{0.790}     & 0.700      \\ 
    Consolidation                & \textbf{0.783}   & 0.619              & 0.780              & 0.780      \\ 
    Infiltration                 & 0.642            & 0.619              & 0.690              & \textbf{0.730} \\ 
    Pneumothorax                 & 0.764            & 0.553              & \textbf{0.860}     & 0.790      \\ 
    Edema                        & 0.880            & 0.680              & \textbf{0.910}     & 0.890      \\ 
    Emphysema                    & \textbf{0.665}   & 0.473              & 0.340              & 0.530      \\ 
    Fibrosis                     & 0.575            & 0.593              & 0.660              & \textbf{0.670} \\ 
    Effusion                     & 0.836            & 0.653              & 0.850              & \textbf{0.870} \\ 
    Pneumonia                    & 0.721            & 0.599              & 0.750              & \textbf{0.770} \\ 
    Pl. Thickening           & 0.675            & 0.435              & 0.420              & \textbf{0.740} \\ 
    Cardiomegaly                 & 0.825            & 0.576              & 0.660              & \textbf{0.940} \\ 
    Nodule                       & 0.494            & 0.549              & \textbf{0.700}     & 0.530      \\ 
    Mass                         & 0.675            & 0.656              & \textbf{0.830}     & 0.750      \\ 
    Hernia                       & 0.591            & 0.404              & 0.830              & \textbf{0.800} \\ 
    No Finding                   & 0.277            & 0.442                & 0.600              & \textbf{0.640} \\ 
    \midrule
    \textbf{Average AUC}         & 0.677            & 0.556              & 0.720              & \textbf{0.742} \\
    \bottomrule
    \end{tabular}
    \label{tab:auc_comparison_moco}
\end{table}

We then integrated MoCo into the image encoder of the CXRCLIP model to create our proposed model, \textbf{MoCoCLIP}. The MoCoCLIP model achieved an average AUC of 0.742, outperforming both CXRCLIP and CheXZero. Table \ref{tab:auc_comparison_moco} presents a comparison of AUC scores for various pathologies across the models.

These enhancements demonstrate that MoCo effectively boosts the model’s discriminative capability, particularly for pathologies that are challenging to detect due to subtle imaging features. However, for some pathologies like \emph{Pneumothorax}, the AUC slightly decreased from 0.860 to 0.850, suggesting that the effectiveness of MoCo may vary depending on the specific pathology or dataset characteristics.

\subsection{Zero-Shot Inference on CheXpert Dataset}
\label{ssec:subhead}
To evaluate the generalization ability of our models, we performed zero-shot inference on the CheXpert test dataset. Table \ref{tab:chexpert} presents the AUC scores for various pathologies across CheXZero, CXRCLIP, and MoCoCLIP.

MoCoCLIP achieved an average AUC of 0.750, outperforming both CheXZero (0.746) and CXRCLIP (0.733). Pleural Effusion achieved the highest AUC of 0.939 with our model. In addition, this study has seen a drop in performance with the MoCo model for some pathologies like Atelectasis and Edema, which shows that integrating MoCo may have mixed effects depending on the specific characteristics of each pathology.

\begin{table}[t]
\centering
\caption{AUC performance comparison of CheXZero, CXRCLIP, and MoCoCLIP on CheXpert test dataset.}
\footnotesize
\begin{tabular}{lccc}
\toprule
\textbf{Pathology}            & \textbf{Chexzero} & \textbf{CXRCLIP} & \textbf{MoCoCLIP} \\ \midrule
Atelectasis                   & 0.788            & 0.798         & 0.676         \\ 
Cardiomegaly                  & 0.893             & 0.867         & 0.862         \\ 
Consolidation                 & 0.891            & 0.846         & 0.819         \\ 
Edema                         & 0.906            & 0.844         & 0.829          \\ 
Enlarged Card.                & 0.894            & 0.812         & 0.661         \\ 
Fracture                      & 0.743            & 0.587         & 0.482         \\ 
Lung Lesion                   & 0.683            & 0.705         & 0.784          \\ 
Lung Opacity                  & 0.916            & 0.818         & 0.723         \\
Pleural Effusion              & 0.930            & 0.936         & 0.939         \\ 
Pleural Other                 & 0.616            & 0.623         & 0.698         \\ 
Pneumonia                     & 0.810            & 0.737         & 0.750         \\ 
Pneumothorax                  & 0.631            & 0.692          & 0.682         \\ 
Support Devices               & 0.671            & 0.797          & 0.758         \\ 
No Finding                    & 0.073            & 0.200         & 0.827         \\ 
\midrule
\textbf{Average}                  & \textbf{0.746}   & \textbf{0.733} & \textbf{0.750}  \\ 
\bottomrule
\end{tabular}
\label{tab:chexpert}
\end{table}

\subsection{Discussion}

The experimental results confirm that \textbf{MoCoCLIP} establishes a new state-of-the-art in zero-shot medical imaging classification. By integrating MoCo into the CLIP framework, the model learns more robust and discriminative features from unlabeled data. This improvement is attributed to MoCo’s momentum encoder and a large queue to store negative samples, which overcomes the limitation of requiring large batch sizes in CLIP for effective contrastive learning. By providing a consistent and diverse set of negative examples, MoCoCLIP mitigates the impact of class imbalance, which is common in medical imaging datasets.

The improvements in AUC scores, particularly for challenging conditions like \emph{Cardiomegaly}, \emph{Infiltration}, and \emph{Pleural Thickening}, highlight the effectiveness of our approach. Nevertheless, the variability in performance across different pathologies suggests that while MoCoCLIP generally improves model performance, it may require pathology-specific adjustments. Further investigation is needed to optimize the model for conditions where performance decreased.

\section{Abalation Study}
\label{sec:majhead}
We show the impact of batch size and various loss functions in the following two ablation studies. 

Table \ref{tab:comparison_batch_16_32} denotes the performance improvement when the batch size is increased from 16 to 32, denoting an average AUC increase of approx 4\%. The larger batch size provides more stable gradient updates, that enhance learning. 
\begin{table}[t]
\centering
\caption{AUC Scores for different pathologies with batch size 16 and batch size 32 where the AUC scores are compared.}
\footnotesize
\begin{tabular}{lcc}
\toprule
\textbf{Pathology} & \textbf{bs=16} & \textbf{bs=32} \\ \midrule

Pneumothorax        & 0.768 & \textbf{0.817} \\ 
Edema               & 0.810 & \textbf{0.880} \\ 
Fibrosis            & 0.573 & \textbf{0.614} \\ 
Effusion            & 0.835 & \textbf{0.860} \\ 
Pneumonia           & 0.671 & \textbf{0.699} \\ 
Cardiomegaly        & 0.887 & \textbf{0.889} \\ 
\midrule
\textbf{Average}    & 0.757 & \textbf{0.793} \\ \bottomrule
\end{tabular}
\label{tab:comparison_batch_16_32}
\end{table}
 
\begin{table}[t]
\centering
\caption{Mean AUC Scores for Different Loss Techniques}
\footnotesize
\begin{tabular}{lr}
\toprule
\textbf{Loss Technique} & \textbf{Mean AUC} \\ \midrule
\textbf{A: MoCo Loss + Image-Text Contrastive Loss} & \textbf{0.742} \\ 
B: Contrastive Loss + Momentum Encoder Text Loss & 0.727 \\ 
C: Contrastive Loss + Momentum Encoder Image Loss & 0.700 \\ 
D: Full Losses Integration & 0.734 \\ \bottomrule
\end{tabular}
\label{tab:mean_auc_comparison_loss}
\end{table}

Additionally, experiments were also conducted to evaluate the impact of different combinations of loss functions on the model’s performance. Table \ref{tab:mean_auc_comparison_loss} presents the mean AUC scores for each loss integration strategy. In our findings, Configuration A achieved the highest mean AUC of 0.744, indicating that combining the MoCo Loss with the Image-Text Contrastive Loss is the most effective strategy for improving model performance. Config D, which integrates all loss terms, resulted in a slightly lower mean AUC of 0.734, suggesting that adding Momentum Encoder Losses for both image and text does not provide additional benefits. Configurations B and C, which include the Momentum Encoder Loss for text and image respectively, showed lower mean AUC scores of 0.727 and 0.700. These results suggest that the most effective approach is to use the MoCo Loss combined with the Image-Text Contrastive Loss without incorporating the additional Momentum Encoder Losses.

\section{Conclusion and Future Work}
\label{sec:conclusion}

In this paper, we introduced \textbf{MoCoCLIP}, an enhanced model for zero-shot learning in medical imaging, achieved by integrating Momentum Contrast (MoCo) into the CLIP framework. Our method addresses challenges associated with unlabeled and class-imbalanced datasets, enabling improved detection of pulmonary and cardio-thoracic pathologies without extensive annotated data.

Experimental results on the NIH CXR14 and CheXpert datasets demonstrate that \textbf{MoCoCLIP} outperforms baseline models, including the state-of-the-art \textbf{CheXZero}, achieving higher AUC scores in classification tasks. The ablation studies further highlight the effectiveness of MoCo integration and provide insights into optimal training configurations. While our approach shows promise, the variability in performance across different pathologies suggests that further optimization is necessary. Moreover, our use of synthetic reports instead of real radiology reports may have limited the model’s ability to fully leverage contextual information. Incorporating real CXR reports could potentially enhance performance and generalization.

\section{COMPLIANCE WITH ETHICAL STANDARDS}
This research study was conducted using public data made available in open access. Ethical approval was not required as confirmed by the license attached with the open-access data.

\section{Acknowledgments}
The authors have no relevant financial or non-financial interests to disclose. The authors declare that there are no conflicts of interest with respect to this publication.

\bibliographystyle{IEEEbib}
\bibliography{main}

\end{document}